\documentclass[a4paper,fleqn]{cas-sc}

\usepackage[authoryear,longnamesfirst]{natbib}

\def\tsc#1{\csdef{#1}{\textsc{\lowercase{#1}}\xspace}}
\tsc{WGM}
\tsc{QE}

\begin{document}
\let\WriteBookmarks\relax
\def\floatpagepagefraction{1}
\def\textpagefraction{.001}

\shorttitle{}    


\title [mode = title]{PhysAttNet: Enhancing Predictive Performance in Industrial and Astrophysical Time Series via Physics-Informed Attention}  








\author[1]{Amal Saadallah}[orcid=0000-0003-2976-7574]
\cormark[1]
\ead{amal.saadallah@ruhr-uni-bochum.de}


\affiliation[1]{organization={ Faculty of Physics and Astronomy, Ruhr University Bochum},
                country={Germany}}

\author[1]{Julia Tjus}[style=chinese]

\author[2]{Petra Wiederkehr}[%
   ]


\affiliation[2]{organization={Virtual Machining Group, TU Dortmund},
                country={Germnay}}

\author[3]{Wolfgang Rhode}

\affiliation[3]{organization={Faculty of Physics},
                country={Germany}}

\cortext[cor1]{Corresponding author}

\begin{abstract}
Accurate and robust time series forecasting is essential in many applied domains involving physical processes, such as manufacturing monitoring and astrophysical event detection. In these settings, predictive models must remain reliable under noise, variability, and measurement uncertainty while capturing temporally localized structures that correspond to physically meaningful events. Convolutional neural networks (CNNs) are widely used for such tasks due to their computational efficiency and strong representational capacity. However, their learned temporal representations often exhibit unstable or physically inconsistent attention patterns, which can reduce robustness, generalization, and interpretability.
This paper introduces PhysAttNet, a physics-informed attention framework for time series forecasting. PhysAttNet augments a lightweight CNN forecaster with an additional attention head whose behavior is guided by domain-informed regularization reflecting structural properties of physical signals. Specifically, three complementary constraints are imposed during training: (i) an alignment regularization that encourages attention to follow smooth peak-centered temporal structures derived from the input signal, (ii) a smoothness regularization that enforces continuous temporal evolution, and (iii) a sparsity regularization that promotes selective focus on compact and informative intervals. These differentiable regularization terms introduce physics-guided inductive bias without requiring annotated explanations or manual supervision.
Experiments based on two very different application scenarios, namely predicting cutting forces during milling and forecasting flares in blazar time series, demonstrate that PhysAttNet improves forecasting accuracy and generalization while enhancing prediction performance on structurally important events. The approach achieves these benefits while maintaining a lightweight and computationally efficient architecture, making it suitable for real-world scientific and industrial time series applications.

\end{abstract}




\maketitle

\section{Introduction}

Forecasting temporal phenomena in scientific and industrial domains requires models that achieve high predictive accuracy while remaining consistent with the underlying dynamics of the system \citep{saadallah2019drift,saadallah2018stability}. In many real-world settings, however, purely data-driven approaches face fundamental limitations. Observational data are often limited, noisy, or highly specific to particular operating conditions, while the processes generating the data are governed by strong physical constraints and structured temporal dependencies \citep{saadallah2022online}. As a result, models that rely solely on statistical correlations may capture short-term patterns but fail to generalize across varying system states or operating regimes \citep{saadallah2022explainable,saadallah2021explainable}.

These challenges arise across a wide range of domains where temporal signals are produced by physical processes. For instance, in manufacturing systems, cutting-force measurements in milling operations exhibit structured patterns driven by tool–material interactions, where continuous changes in the cutting edge (i.e., tool wear) are reflected in a gradual change in the time signal and chipping at the cutting edge or process instabilities lead to abrupt changes \citep{finkeldey2020real,saadallah2018stability}. In high-energy astrophysics, blazar light curves recorded by gamma-ray telescopes contain flare events characterized by sparse and most of the time irregular rise and decay phases that reflect particle acceleration processes within relativistic jets \citep{georgoulis2021flare}. 
While these domains differ substantially in scale and physical interpretation, they share key characteristics: important events occur in temporally localized regions, where the signal exhibits continuous temporal evolution around peak-centered structures, and meaningful patterns must be distinguished from noise and transient fluctuations.
Such properties highlight the need for machine learning models that not only learn from data but also incorporate structural knowledge about the underlying system. Embedding domain-informed constraints into deep learning models can improve robustness, guide the learning process toward physically plausible representations, and enhance generalization when data are sparse or noisy \citep{nasir2025embedding}. This perspective is increasingly reflected in hybrid approaches that combine data-driven learning with domain knowledge, including frameworks such as Physics-Informed Neural Networks \citep{luo2025physics,nasir2025embedding} and related constraint-aware architectures \citep{baez2024guaranteeing}.

Despite these advances, many deep learning models for time series forecasting remain largely unconstrained. In particular, convolutional neural networks (CNNs) and related architectures often achieve strong predictive performance but lack mechanisms to ensure that their internal attention aligns with known physical characteristics of the input signals \citep{vonder2023analysis}. Attention maps derived from gradient-based saliency methods \citep{selvaraju2016grad} frequently highlight diffuse or noisy regions \citep{adebayo2018sanity,saadallah2021explainable,saadallah2022explainable}, resulting in explanations that may be inconsistent with domain knowledge. While recent attention-based architectures \citep{li2025novel,gao2021interpretable,tamayo2025your} attempt to address this limitation, they often introduce substantial architectural complexity and computational overhead.

To address these challenges, we propose PhysAttNet, a physics-informed attention-based convolutional neural network for time series forecasting. Instead of enforcing explicit physical equations, PhysAttNet incorporates shared structural assumptions about temporal signals, localized events, continuous evolution within peak-centered temporal windows, and selective relevance of past observations, directly into the training objective. This is achieved through attention regularization that guides the model to focus on coherent temporal regions where meaningful dynamics occur, while suppressing diffuse or noisy responses.
By encouraging the model to concentrate on structured peak-centered regions, the proposed approach not only produces physically consistent explanations but also improves predictive performance on the events that are most relevant to domain experts, such as force peaks in milling processes or flare maxima in blazar light curves.

To ensure that the resulting attention maps remain interpretable and physically meaningful, PhysAttNet integrates three complementary regularization terms. The first is an alignment regularization, which encourages agreement between the model-generated attention and Gaussian-shaped target profiles centered on relevant temporal regions. The second is a continuity regularization, designed to suppress abrupt variations and enforce continuous temporal evolution within these regions. The third is a sparsity regularization, which limits the spread of attention and promotes focus on compact, informative intervals. Together, these components produce attention maps that are localized, continuous, and aligned with domain knowledge. Importantly, attention supervision requires only minimal architectural modification, implemented through a lightweight additional attention head. This design maintains computational efficiency and makes the model suitable for real-time or resource-constrained deployment.

The contributions of this work are threefold:
\begin{itemize}
    \item A novel physics-aligned attention-supervised CNN architecture (PhysAttNet) that embeds domain-driven temporal structure directly into its learning objective.
    \item A unified supervised-attention framework combining alignment, smoothness, and sparsity regularizers to produce coherent and physically consistent temporal explanations.
    \item Extensive evaluation on two distinct real-world forecasting problems, namely milling cutting-force prediction and blazar flare detection. The results demonstrate improved predictive accuracy, and stronger alignment with expert-understood temporal patterns.
\end{itemize}

\section{Related Works}

\subsection{Physics-Informed Deep Learning for Time Series Forecasting}

A substantial line of research has introduced Physics-Informed Neural Networks (PINNs) and domain-constrained deep models that embed physical laws, structural priors, or scientific constraints directly into the learning process \citep{luo2025physics,nasir2021review}. Classical PINN formulations incorporate differential equation residuals into the loss to ensure physically consistent solutions \citep{luo2025physics}, while recent adaptations impose monotonicity, energy conservation, or domain-specific invariants to guide representation learning \citep{chu2024structure,baez2024guaranteeing}. These approaches have shown that embedding scientific knowledge improves generalization in sparse-data regimes and stabilizes predictions under distribution shift.

However, most physics-informed architectures focus on continuous dynamics, static PDE formulations, or simulation-to-real transfer, and thus do not directly address convolutional forecasting of discrete temporal signals. Furthermore, prior work rarely integrates physics constraints with deep temporal attention mechanisms or explanation-guided learning. Compared to these methods, PhysAttNet introduces a lightweight but expressive physics-inspired attention module that enforces domain priors through structured masks and continuity constraints directly on temporal relevance scores, something not explored in PINNs or physics-based CNNs. Unlike classical PINNs, our method does not require differential equations or strong priors. Instead, it generalizes physics-inspired regularity into a flexible attention mechanism suitable for noisy real-world forecasting.

\subsection{Attention Mechanisms for Time Series Forecasting}

Attention-based architectures have reshaped time series forecasting, from global models such as the Temporal Fusion Transformer (TFT) \citep{nazir2023forecasting} and Informer \citep{gong2023short} to lightweight hybrid models combining CNNs or RNNs with temporal self-attention \citep{li2025novel}. These methods enable long-range dependency modeling, variable selection, and dynamic weighting of historical segments. Improvements in interpretability have been reported \citep{gao2021interpretable,tutek2022toward}, yet several studies indicate that attention weights can remain unstable, non-monotonic, or misaligned with true causal signals, particularly under noise or distribution shift \citep{jain2019attention,tamayo2025your}.

In contrast with these transformer-style approaches, PhysAttNet employs a physics-inspired attention block designed specifically for 1D CNN forecasting. Instead of unconstrained attention weights, we incorporate structural priors to stabilize temporal relevance and reflect physical constraints of the underlying system. This distinguishes our method from generic attention mechanisms. 

\subsection{Explanation-Guided and Explanatory Interactive Learning (XIL)}
In time series analysis, common post-hoc explanation methods include gradient-based saliency \citep{saadallah2021explainable,saadallah2022explainable}, perturbation-based attribution \citep{schlegel2023deep}, and Shapley-value approaches \citep{jakobs2023explainable}, which identify input regions or temporal segments that contribute most strongly to model predictions. While these techniques provide useful insights, they do not influence the training process and therefore cannot prevent models from learning spurious or physically implausible decision strategies.

Beyond post-hoc interpretation, recent advances in explanation-guided training and explanatory interactive learning (XIL) demonstrate that model performance and reliability can be improved by explicitly constraining explanations during learning \citep{teso2019explanatory}. Representative approaches include Grad-CAM–guided losses that penalize misaligned saliency maps \citep{selvaraju2021casting}, prototype-based alignment methods that enforce reliance on interpretable reference patterns \citep{huang2025protopgtn,brenner2024concept}, and human-in-the-loop XIL frameworks that iteratively refine models using expert feedback on explanations \citep{teso2019explanatory}. In the time series domain, methods such as RioT  \citep{kraus2025right} further extend this paradigm by constraining temporal or frequency-domain attributions to align with domain-informed relevance priors, reducing spurious correlations and improving robustness under noise and distribution shifts.
Despite these advances, existing XIL approaches remain limited in scope. Most focus on computer vision \citep{selvaraju2021casting}, multi-modal \citep{teso2019explanatory}, or classification \citep{huang2025protopgtn} tasks and rely on additional supervision in the form of annotated explanations \citep{brenner2024concept}, human rationales \citep{herrewijnen2024human}, or manually specified relevance masks \citep{brenner2024concept}. Applications to time series forecasting are comparatively rare, and the integration of explanation-guided losses with structured attention mechanisms or physics-informed constraints is largely unexplored. Moreover, many approaches operate as corrective mechanisms on top of complex architectures, increasing training complexity and limiting deployability.


In contrast to prior work, PhysAttNet integrates explanation-aligned behavior directly into the model architecture and training objective by combining structured attention with physics-informed regularization. This enables the model to learn temporally coherent, localized, and physically plausible attention patterns without requiring explicit explanation annotations or human feedback, bridging the gap between explainability, robustness, and practical time series forecasting.

\section{Application Scenarios and Domain Assumptions}
To demonstrate the generality of PhysAttNet, two application scenarios from distinct scientific domains are considered, specifically high-energy astrophysics and manufacturing process monitoring. Although these domains differ substantially in scale, measurement characteristics, and physical interpretation, both produce time series governed by underlying physical processes that generate structured events embedded in noisy background signals. In particular, important phenomena appear as temporally localized peaks with continuous evolution within short time windows. These shared structural properties motivate forecasting models that selectively focus on informative temporal segments while remaining robust to spurious variations in the signal.

\subsection{Astrophysical Flare Forecasting from Fermi-LAT Observations}

\begin{figure}[t]
\centering
\includegraphics[width=0.85\linewidth]{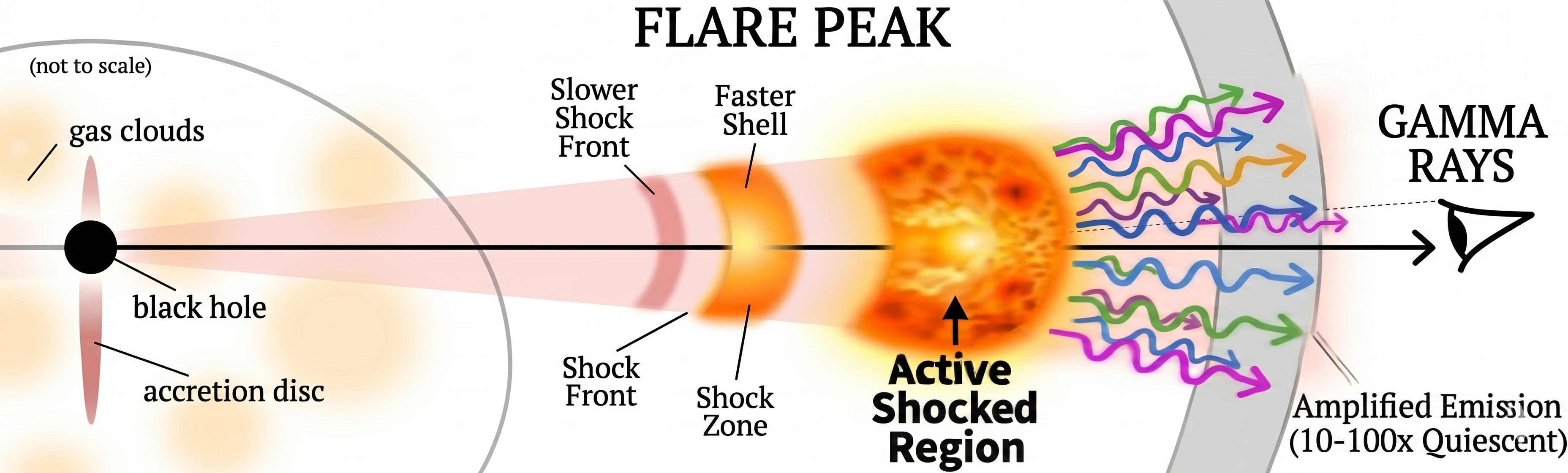}
\caption{Conceptual illustration of the physical processes leading to gamma-ray flares in blazar jets. Shock interactions and particle acceleration within the relativistic jet can produce localized bursts of high-energy emission that appear as flares in the observed gamma-ray time series.}
\label{fig:blazar_process}
\end{figure}

The first use case considers the forecasting of gamma-ray flares from blazars observed by the \textit{Fermi} Large Area Telescope (Fermi-LAT) \citep{abdo2010gamma}. Blazars are a class of active galactic nuclei whose relativistic jets are oriented close to the observer’s line of sight, resulting in strongly Doppler-boosted and highly variable gamma-ray emission. As illustrated in Figure~\ref{fig:blazar_process}, energetic processes within the jet, such as shock acceleration and particle interactions, can produce localized bursts of radiation that manifest as flares in the observed light curves.

\begin{figure}[t]
\centering
\includegraphics[width=0.99\linewidth]{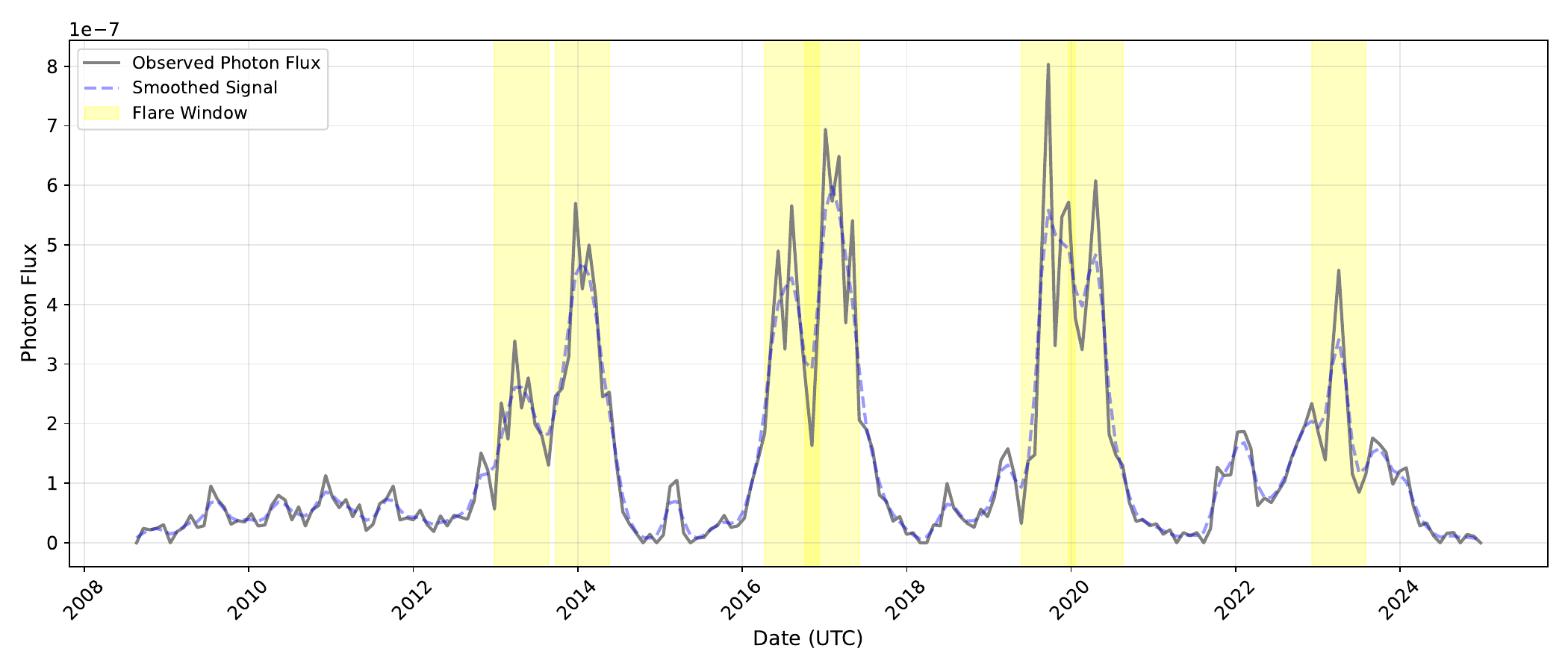}
\caption{Photon flux of the bright $\gamma$-ray source 4FGL J1048.4+7143 observed by Fermi-LAT as a function of time, illustrating the occurrence of localized flare events in the light curve.}
\label{fig:blazar_timeseries}
\end{figure}

We analyze observations from 13 blazars identified in the Fourth Fermi-LAT Source Catalog (4FGL). These sources were selected to provide a representative set of blazar light curves with sufficient temporal coverage, variability, and data quality for consistent forecasting evaluation across multiple sources and time resolutions. The selection spans sources exhibiting diverse flare activity patterns while remaining computationally manageable for systematic benchmarking. For each source, we use photon flux measurements in the energy range 0.1–100 GeV as the target time series. The data are available at daily, weekly, and monthly temporal resolutions, allowing the study of forecasting behavior across multiple time scales. Although each observation includes additional metadata such as test statistics, photon index estimates, and uncertainty measures, the forecasting task focuses on photon flux as the primary physical observable. Figure~\ref{fig:blazar_timeseries} shows the photon flux of the bright $\gamma$-ray source 4FGL J1048.4+7143 as a function of time.

From a physical perspective, gamma-ray flares manifest as localized bursts of emission caused by particle acceleration and energy dissipation processes within the relativistic jet. These flares typically exhibit gradual rise and decay phases, forming peak-centered temporal structures rather than abrupt discontinuities. Outside flare intervals, the photon flux evolves more slowly, reflecting quiescent emission states.

This combination of localized high-amplitude events and continuous temporal evolution motivates forecasting models that emphasize salient temporal regions while maintaining coherent temporal dynamics. In particular, models should selectively attend to flare-centered intervals that drive future emission behavior while avoiding overreaction to short-lived statistical fluctuations. These considerations directly motivate the attention alignment, temporal coherence, and localization constraints incorporated in our framework.

\subsection{Cutting Force Forecasting in Machining Processes}

\begin{figure}[t]
\centering
\includegraphics[width=0.85\linewidth]{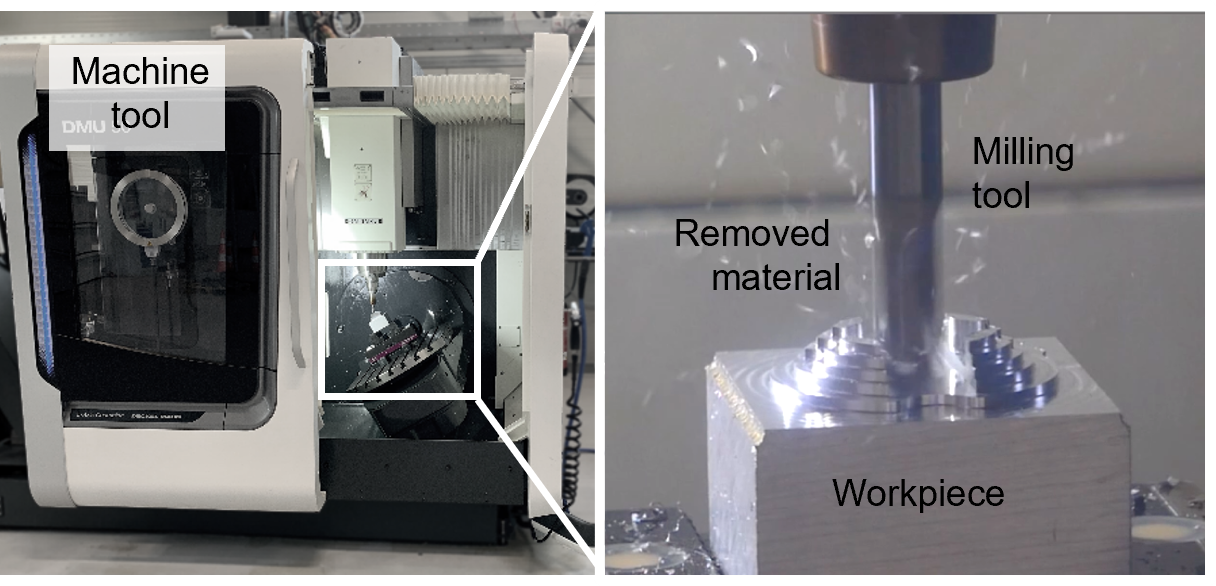}
\caption{Illustration of the NC milling process. A rotating milling tool removes material from the workpiece, generating cutting forces that are measured during machining.}
\label{fig:milling_process}
\end{figure}

The second use case considers high-frequency cutting force measurements collected during numerical control (NC) milling operations. NC milling is a subtractive manufacturing process in which a rotating cutting tool removes material from a workpiece to produce a desired geometry. This process is widely used in production engineering, for example in tool and mold manufacturing and in the fabrication of high-precision components such as turbine blades in the aerospace industry \citep{wiederkehr2016virtual,finkeldey2020real,saadallah2018stability}. Figure~\ref{fig:milling_process} illustrates the machining setup, including the milling tool, the workpiece, and the material removal process during cutting.

\begin{figure}[t]
\centering
\includegraphics[width=0.9\linewidth, height=5.2cm]{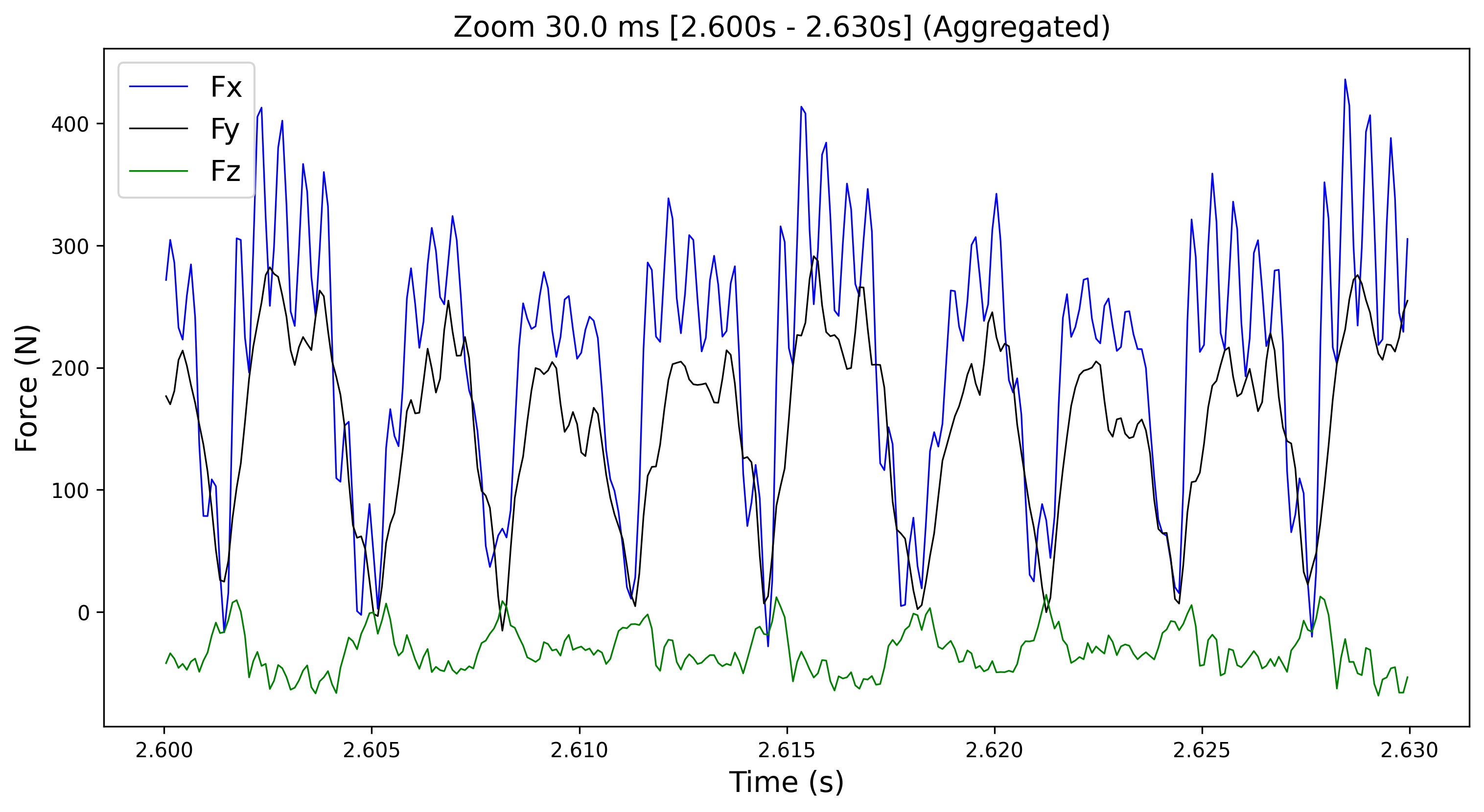}
\caption{Progression of cutting forces on zoomed time intervals.}
\label{fig:force_aggregation}
\end{figure}

We analyze 274 multivariate time series recorded under varying process conditions, including cutting speed, feed rate, and experimental configurations. Each time series contains high-frequency measurements of cutting forces along three axes ($F_x$, $F_y$, $F_z$). The original signals are recorded at a sampling frequency of 200 kHz (5 µs resolution). For forecasting purposes, the signals are aggregated to a temporal resolution of 0.1 ms. The objective of this aggregation is not to resolve individual tooth engagement events of the milling cutter, but to capture the longer-term evolution of cutting forces, which is more relevant for monitoring gradual changes in the process, such as tool wear or process instability. 
Figure~\ref{fig:force_aggregation} shows an example of the resulting force signals compared to the original measurements, illustrating that the aggregated series preserves the characteristic force patterns and the dominant peak-centered temporal dynamics relevant for forecasting future cutting behavior, while reducing high-frequency noise.

Cutting forces arise from the physical interaction between the cutting tool and the workpiece, governed by material deformation, friction, and chip formation. These interactions generate force signals with characteristic temporal structure. Forces typically increase as the tool engages the material, reach localized maxima during stable cutting, and decrease as the tool disengages or process conditions change. Tool wear and degradation are reflected not by isolated impulses but by systematic changes in the magnitude, duration, and shape of these peak-centered force responses over time.
Although high-frequency oscillations are present due to vibrations and measurement noise, the underlying force evolution is still constrained by mechanical inertia and material properties, resulting in predominantly continuous and structured dynamics. At the same time, abrupt variations can occur in practice, for example, due to chatter, transient tool–workpiece interactions, or changes in engagement conditions. In the controlled experiments considered here, however, the tool follows uniform paths with largely consistent engagement conditions, meaning that most informative patterns appear as structured peak-centered force responses. Consequently, isolated spikes or rapidly fluctuating patterns are more likely to reflect sensor noise or short-lived disturbances rather than sustained changes in the machining process.

Effective forecasting in this setting, therefore, requires models that emphasize coherent, localized force patterns while remaining robust to transient fluctuations. In particular, attention mechanisms should focus on peak-centered temporal segments that drive future force evolution, rather than reacting to short-lived noise. These considerations motivate the use of attention constraints that promote temporal coherence and selective focus on compact, informative intervals of the force signal.

\subsection{Shared Domain Assumptions}

Although the two application scenarios originate from fundamentally different scientific domains, namely high-energy astrophysics and manufacturing process monitoring, they share several structural characteristics at the level of the observed time series. The differences between the domains are substantial. In astrophysics, gamma-ray observations capture radiation emitted by relativistic jets from supermassive black holes located billions of light-years away, with photon flux measurements collected at daily to monthly temporal resolutions. In contrast, machining process monitoring involves force signals measured directly on industrial equipment at millisecond or microsecond sampling rates during tool and workpiece interactions. These differences reflect distinct physical scales, measurement instruments, and temporal resolutions.

Despite these domain-specific characteristics, the resulting time series exhibit comparable structural properties that can be exploited for forecasting. First, important events tend to appear as temporally localized peak-centered patterns, such as force peaks during tool–workpiece engagement in machining or flare maxima in gamma-ray light curves. Second, although the signals may contain noise and short-lived fluctuations, the dominant physical dynamics evolve over continuous temporal intervals rather than single isolated time points. 
These shared characteristics motivate a unified modeling approach in which attention mechanisms are guided by domain-informed regularization rather than relying purely on data-driven learning. The proposed PhysAttNet architecture incorporates these assumptions through attention alignment, temporal coherence, and localization constraints, enabling the model to focus on physically meaningful temporal regions while maintaining a lightweight and computationally efficient design applicable across domains.

\section{Methodology}

Building on the application scenarios introduced above, this section presents \emph{PhysAttNet}, a physics-informed attention-based CNN architecture for time series forecasting. While the considered domains—cutting-force monitoring in machining and astrophysical flare analysis—differ substantially in their physical scale and measurement conditions, both produce time series in which predictive information is often concentrated within temporally localized events embedded in broader system dynamics.
These events may appear as force responses associated with tool–workpiece engagement in milling processes or as flare activity in astrophysical observations. Importantly, the framework does not assume strictly smooth dynamics. Hence, rapid transitions may occur, for example due to chatter in machining processes or sudden flare onset in astrophysical sources. Nevertheless, informative signal segments typically form coherent temporal regions rather than isolated fluctuations.
Rather than imposing explicit physical equations, requiring expert-annotated explanations, or manually labeling relevant temporal regions, PhysAttNet incorporates these domain-informed assumptions through differentiable attention regularization terms derived directly from the observed signals. These regularizers guide the model toward physically plausible temporal focus patterns while preserving the flexibility of data-driven learning.

\subsection{Problem Formulation}

Let $\{x_t\}_{t=1}^{T}$ denote a univariate ($d=1$) or multivariate ($d>1$) time series with observations $x_t \in \mathbb{R}^d$. Given a look-back window of length $L$, the forecasting task is to predict a future value $x_{t+h}$ at horizon $h$. Using the input sequence
\[
\mathbf{x}_t = [x_{t-L+1}, \ldots, x_t],
\]
the goal is to learn a forecasting function $f_\theta$ such that
\[
x_{t+H}\hat{x}_{t+h} = f_\theta(\mathbf{x}_t),
\]
where $\theta$ denotes the model parameters.
PhysAttNet decomposes this mapping into three components, namely
(i) a lightweight CNN backbone for temporal feature extraction,
(ii) a temporal attention mechanism that selectively aggregates informative features, and
(iii) physics-informed regularization terms that constrain the learned attention patterns during training.

\subsection{CNN Backbone for Temporal Feature Extraction}

The backbone network consists of stacked one-dimensional convolutional layers operating on the time-delay embedded input $\mathbf{x}_t$. These convolutional filters capture short- to mid-range temporal structures such as gradual trends, oscillatory behavior, and transient peaks.

In machining signals, such patterns may correspond to tool–workpiece engagement dynamics, force responses generated by cutting edges, or disturbances such as chatter. In astrophysical observations, similar structures arise from flare rise and decay phases or variations in emission intensity.

The architecture is intentionally lightweight and avoids deep transformer-based structures. This design reflects the practical constraints of operational monitoring systems and ensures that improvements in predictive performance arise from the proposed attention regularization rather than from architectural complexity alone. 
\subsection{Physics-Informed Temporal Attention}

Let $\mathbf{X} \in \mathbb{R}^{L \times d}$ denote an input time series segment of length $L$ with $d$ input channels. After a sequence of one-dimensional convolutional and pooling layers, the network produces an intermediate feature representation:
\[\mathbf{H} = [\mathbf{H}_1, \ldots, \mathbf{H}_{T'}] \in \mathbb{R}^{T' \times C}\]
where $T' < L$ is the reduced temporal dimension induced by down-sampling operations and $C$ is the number of learned feature channels. Each vector $\mathbf{H}_t \in \mathbb{R}^{C}$ encodes local temporal patterns extracted at time index $t$.
To enable selective aggregation of temporally informative features, PhysAttNet employs a temporal attention mechanism that assigns a relevance weight to each time step in $\mathbf{H}$. Specifically, an attention score function $g(\cdot)$ maps each feature vector to a scalar score:
\[e_t = g(\mathbf{H}_t), \quad t = 1, \ldots, T'\]
where $g(\cdot)$ is implemented as a lightweight $1 \times 1$ convolution followed by a nonlinearity. The raw scores $\{e_t\}$ are normalized via a softmax operation to obtain attention weights:
\[
\alpha_t = \frac{\exp(e_t)}{\sum_{k=1}^{T'} \exp(e_k)}, \quad \alpha_t \geq 0, \quad \sum_{t=1}^{T'} \alpha_t = 1.
\]

The attention weights define a convex combination over temporal features, yielding the attended representation:
\[\tilde{\mathbf{H}} = \sum_{t=1}^{T'} \alpha_t \mathbf{H}_t \in \mathbb{R}^{C}\]
which is subsequently passed to a regression head to produce the final forecast.

In contrast to purely data-driven attention mechanisms, which may assign high relevance to spurious or noisy temporal patterns, PhysAttNet constrains the attention vector $\boldsymbol{\alpha}$ using domain-informed regularization terms. These constraints encode soft physical priors commonly shared across physical time series, including  the presence of localized event-driven dynamics, temporal continuity, and selective relevance of past observations.
Rather than enforcing explicit physical equations, these priors are incorporated through differentiable penalties applied directly to $\boldsymbol{\alpha}$ during training. As a result, the attention mechanism is guided toward smooth, concentrated, and physically plausible temporal focus patterns, while remaining fully learnable and compatible with standard gradient-based optimization.


\subsubsection{Attention Alignment Regularization}

Many physical time series exhibit localized events with peak-centered structures, such as gamma-ray flares or force transients during tool engagement. To encode this prior, we construct a target attention profile $\boldsymbol{\alpha}^\star$ using Gaussian kernels centered at salient peaks detected in the input signal.
The alignment loss is defined as:
\[
\mathcal{L}_{\text{align}} =
\frac{1}{T'} \sum_{t=1}^{T'} (\alpha_t - \alpha_t^{\star})^2.
\]
This regularizer encourages the model to allocate attention to temporally localized, physically plausible regions without requiring manual annotation of explanations. The Gaussian targets provide soft guidance rather than hard constraints, allowing flexibility when event shapes vary across datasets or scales.



\subsubsection{Attention Smoothness Regularization}

Physical systems often exhibit structured temporal evolution due to underlying system dynamics, such as inertia in mechanical processes or energy dissipation in astrophysical environments. While abrupt transitions may occur, for example,  during chatter events in machining or rapid flare onset in astrophysical sources, relevant signal segments typically form coherent temporal regions rather than isolated single-point in time fluctuations.

To encourage temporally coherent attention patterns, we introduce the smoothness penalty
\[
\mathcal{L}_{\text{smooth}} =
\frac{1}{T'-1} \sum_{t=1}^{T'-1} (\alpha_{t+1} - \alpha_t)^2.
\]

This regularizer discourages rapid oscillations in attention weights that may arise from noise or spurious correlations while allowing the model to adapt to genuine signal transitions.

\subsubsection{Attention Sparsity Regularization}

In many forecasting scenarios, only a small subset of past observations meaningfully influences future behavior. For instance, flare precursors or force transients carry more predictive information than background fluctuations.
To encourage selective focus, we apply an $\ell_1$ sparsity penalty:
\[
\mathcal{L}_{\text{sparse}} =
\frac{1}{T'} \sum_{t=1}^{T'} |\alpha_t|.
\]

 Importantly, this sparsity constraint is not contradictory to the smoothness regularization introduced above. While sparsity promotes \emph{where} attention should be concentrated by discouraging diffuse relevance across time, smoothness governs \emph{how} attention evolves within relevant regions by preventing rapid, noisy oscillations. Together, the two regularizers encourage attention patterns that are both compact and temporally coherent, aligning with the physical expectation that informative events are localized in time yet evolve continuously. This complementary interaction yields interpretable focus patterns and improves robustness in peak-driven and noisy environments.

 \subsection{Training Objective and Computational Considerations}

PhysAttNet is trained by jointly optimizing forecasting accuracy and physics-informed attention constraints. The overall objective is defined as:
\[
\mathcal{L} =
\mathcal{L}_{\text{pred}}
+ \lambda_{\text{align}} \mathcal{L}_{\text{align}}
+ \lambda_{\text{smooth}} \mathcal{L}_{\text{smooth}}
+ \lambda_{\text{sparse}} \mathcal{L}_{\text{sparse}},
\]
where $\mathcal{L}_{\text{pred}}$ denotes the mean squared error between predicted and observed values. The regularization terms $\mathcal{L}_{\text{align}}$, $\mathcal{L}_{\text{smooth}}$, and $\mathcal{L}_{\text{sparse}}$ encode domain-informed assumptions on temporal relevance, continuity, and selectivity of attention, respectively. The weighting coefficients $\lambda_{\text{align}}$, $\lambda_{\text{smooth}}$, and $\lambda_{\text{sparse}}$ control the trade-off between predictive performance and physical plausibility and are selected via validation.
From a computational perspective, PhysAttNet introduces only marginal overhead compared to standard CNN-based forecasters. The attention mechanism operates on temporally downsampled feature representations, and all regularization terms are fully differentiable and inexpensive to compute. Consequently, training and inference remain efficient, making the model suitable for large-scale, high-frequency, or resource-constrained forecasting scenarios while providing improved robustness and interpretability.

\subsection{Adaptable Physics-Informed Attention Constraints}

While PhysAttNet is motivated by application scenarios characterized by localized, continuous events, the proposed framework is intentionally designed to remain flexible and extensible across different types of physical temporal dynamics. Crucially, the physics-informed priors are introduced through modular regularization terms acting on the attention vector $\boldsymbol{\alpha}$, rather than being hard-coded into the network architecture or loss function.
This design allows the attention constraints to be readily adapted to alternative domain assumptions like applications where salient information is associated with local minima rather than peaks, such as pressure drops, force relief during tool disengagement, pressure drops preceding mechanical faults, voltage sags in power systems, or quiescent phases separating energetic astrophysical events. In such cases, regions of reduced signal magnitude carry stronger predictive relevance than high-amplitude excursions. The target attention profile can be inverted or redefined to emphasize trough-centered structures. For example, to encode this alternative notion of relevance, an inverted target attention profile can be defined as:
\[
\alpha_t^{\star,\text{inv}} = 1 - \alpha_t^{\star},
\]
which assigns higher importance to trough-centered temporal regions while down-weighting peak-dominated intervals. 

Similarly, in settings where events are sparse, intermittent, or inherently discontinuous, the temporal smoothness regularization can be relaxed, replaced, or complemented by alternative constraints that encourage abrupt transitions or piecewise-constant attention behavior.

More generally, the attention alignment loss $\mathcal{L}_{\text{align}}$ can be constructed from arbitrary, signal-derived relevance profiles, while smoothness and sparsity penalties can be adjusted to reflect domain-specific notions of temporal coherence and selectivity. As these regularizers are fully differentiable and decoupled from the forecasting objective, they can be reweighted, removed, or augmented without modifying the underlying CNN or attention mechanism.
This modular formulation positions PhysAttNet as a general framework for incorporating soft, domain-informed temporal priors into attention-based forecasting models. Rather than committing to a single physical interpretation, the method provides a principled interface through which diverse assumptions about temporal relevance can be encoded, tested, and refined in a data-driven manner.

\section{Experiments}

\subsection{Experimental Design}

We evaluate PhysAttNet against a diverse set of statistical, machine learning, and deep learning baselines to assess its effectiveness in physics-driven forecasting scenarios.
All models are trained and evaluated using identical train--validation--test splits and comparable input representations to ensure fair comparison.
Hyperparameters are selected via validation, and all neural models are trained using the mean squared error objective.

\subsubsection{Application-Specific Experimental Setups}

\paragraph{Blazar Flare Forecasting}

The dataset comprises 13 blazars exhibiting intermittent flare activity.
Photon flux in the energy range $[0.1, 100]$~GeV is used as the target variable.
Each forecasting instance uses a look-back window of $L = 24$ months to predict photon flux $H = 6$ months ahead.
This setting reflects a realistic astrophysical use case where early anticipation of flare-prone periods is critical for observation planning and scientific analysis.

\paragraph{Cutting Force Forecasting in Machining Processes}

We analyze sensor-based time-series data obtained from 274 milling experiments with different process parameter settings. Each experiment produces a multivariate force signal measured along the three spatial axes ($F_x$, $F_y$, $F_z$) at microsecond temporal resolution.
To enable proactive process monitoring, raw signals are aggregated to $0.1$ milliseconds using sliding windows that preserve transient force events while suppressing high-frequency noise (see Figure \ref{fig:force_aggregation}).
Forecasting horizons are set to $10$ milliseconds such that predicted force patterns precede significant transients, enabling early detection of tool wear, process instabilities, or abnormal cutting conditions. The cutting forces are normalized across all the files to bring them to the same scale.

\subsubsection{Baselines}
    


We consider a diverse set of forecasting baselines spanning classical statistical models, machine learning approaches, and deep neural architectures. Classical methods include Autoregressive Integrated Moving Average models (ARIMA) \citep{makridakis1997arma}, with model orders selected using the Akaike information criterion, and exponential smoothing (ExpSmooth) models \citep{billah2006exponential} that explicitly capture trend and seasonal components. As non-linear machine learning baselines, we evaluate regression decision tree \citep{spiliotis2022decision} and random forest \citep{breiman2001random} trained on time-delay embedded inputs, enabling the modeling of non-linear temporal dependencies without explicit sequence modeling.

For deep learning baselines, we include a convolutional neural network (CNN) forecaster that shares the same backbone architecture as PhysAttNet but is trained without attention or physics-informed regularization, serving as a controlled ablation. We further consider a hybrid CNN–LSTM \citep{alhussein2020hybrid} that combines convolutional feature extraction with recurrent temporal modeling. In addition, we compare against representative state-of-the-art deep forecasting models, including DeepAR \citep{salinas2020deepar}, a probabilistic autoregressive recurrent network; N-BEATS \citep{oreshkin2019n}, a deep feed-forward architecture based on residual backward and forward blocks; and a Transformer-based model \citep{l2022transformer} employing self-attention mechanisms adapted for time series forecasting.

\subsubsection{Evaluation Metrics}

\paragraph{Root Mean Squared Error (RMSE).}
Overall forecasting accuracy is measured using RMSE, which captures average predictive performance across the entire time series.





\paragraph{Event-Weighted Evaluation Metrics}

To emphasize forecasting accuracy in domain-relevant regions, we adopt an \emph{event-weighted RMSE} (EW-RMSE) that up-weights errors occurring during physically significant events \citep{hyndman2006another,laptev2015generic} such as cutting-force transients or astrophysical flares. We define a weight sequence $\{w_t\}_{t=1}^T$ and compute the weighted RMSE as:
\[EwRMSE = \sqrt{
\frac{1}{T} \sum_{t=1}^{T} w_t \, (y_t - \hat{y}_t)^2}\]

This formulation would inflate the metric according to the absolute magnitude of weights and is less interpretable when comparing across datasets with different numbers of events.  Therefore, we normalizes by the sum of weights, ensuring that the overall metric remains on a scale comparable to standard RMSE while giving relative emphasis to important events: 
\[wRMSE = 
\sqrt{
\frac{\sum_{t=1}^{T} w_t \, (y_t - \hat{y}_t)^2}{\sum_{t=1}^{T} w_t}
}\]

We define the weight $w_t$ each time step $t$ as: 
\[
w_t = 1 + \gamma \cdot \mathbb{I}(y_t > \tau),
\]
where $\mathbb{I}(\cdot)$ is the indicator function, $\tau$ denotes a percentile-based threshold identifying events, and $\gamma > 0$ controls the relative emphasis of the event regions. Percentile-based thresholds are distribution-free and robust to heavy tails and non-stationarity, which are common in both astrophysical and manufacturing signals.

\paragraph{\textbf{Astrophysical flares.}} We use monthly-averaged Fermi-LAT photon flux for 13 Blazars. Flare periods are identified using the $95-th$ percentile of photon flux: 
\[\tau_\text{flare} = \text{Percentile}_{95}(y_t)\]
and we set $\gamma_\text{flare} = 2$, meaning that errors on flare months count two times as much as baseline months. This choice balances sensitivity to rare flares with overall interpretability of RMSE, without being dominated by a few extreme peaks.


 \paragraph{\textbf{Cutting-force transients.}}
For the milling experiments, transient events are defined based on the \emph{resultant cutting force}, which summarizes the combined mechanical load acting on the tool. The resultant force is computed as 
\[
F_{\text{res}} = \sqrt{F_x^2 + F_y^2 + F_z^2},
\]
where $F_x$, $F_y$, and $F_z$ denote the cutting-force components measured along the three axes. We adopt $F_{\text{res}}$ because it provides a physically meaningful scalar representation of the instantaneous cutting load and is widely used in machining analysis and tool-condition monitoring. Transient events are then defined as time points where the resultant force exceeds the 90-th percentile:
\[
\tau_\text{force} = \text{Percentile}_{90}(F_{\text{res}}).
\]
We set $\gamma_\text{force} = 3$, giving increased weight to high-magnitude force transients that are most relevant for proactive tool-wear monitoring. The chosen $\gamma$ value provides a moderate emphasis on rare but critical events, allowing the wRMSE to capture performance in these regions while keeping the metric comparable to standard RMSE.
 
\subsection{Results}

Table~\ref{tab:combined_results} summarizes forecasting performance across both application domains. The metrics are aggregated across the different sources and process configurations for the Blazars forecasting and the milling use cases, respectively. 
In the cutting-force task, deep learning models consistently outperform classical and tree-based baselines, highlighting the importance of modeling structured temporal dependencies in high-frequency physical signals. Transformer and N-BEATS achieve the lowest RMSE values, while PhysAttNet delivers comparable performance across both RMSE and weighted RMSE metrics. Importantly, PhysAttNet attains this competitive accuracy with a substantially simpler architecture, built upon a lightweight CNN augmented by physics-informed attention regularization. This makes it computationally more efficient and structurally more grounded in the underlying process dynamics than large sequence models such as Transformers or multi-block N-BEATS.
In the blazar time series task, PhysAttNet achieves the best overall performance in both RMSE and event-weighted RMSE. While standard CNN and DeepAR models perform strongly, PhysAttNet consistently yields lower mean error and reduced variability, with particularly clear improvements under the weighted metric. This suggests enhanced modeling of high-activity intervals that dominate predictive difficulty in astrophysical light curves.

Overall, the results demonstrate that physics-informed attention regularization enables performance competitive with state-of-the-art deep architectures while retaining architectural simplicity and domain grounding, supporting its suitability for real-world scientific and industrial forecasting applications.

\begin{table}[t]
\centering
\caption{Mean forecasting performance ($\pm$ standard deviation) across the two application domains.}
\label{tab:combined_results}
\resizebox{0.95\textwidth}{!}{
\begin{tabular}{lcc|cc}
\toprule
& \multicolumn{2}{c|}{\textbf{Cutting Forces}} 
& \multicolumn{2}{c}{\textbf{Blazar Time Series}} \\
\cmidrule(lr){2-3} \cmidrule(lr){4-5}
\textbf{Model} 
& \textbf{RMSE} & \textbf{wRMSE} 
& \textbf{RMSE} & \textbf{wRMSE} \\
\midrule
ARIMA         
& $0.881 \pm 0.056$ & $0.857 \pm 0.078$ 
& $9.31\mathrm{e}{-08} \pm 2.92\mathrm{e}{-08}$ 
& $1.08\mathrm{e}{-07} \pm 3.10\mathrm{e}{-08}$ \\

ExpSmooth     
& $1.869 \pm 2.437$ & $1.068 \pm 25525$ 
& $9.45\mathrm{e}{-08} \pm 2.86\mathrm{e}{-08}$ 
& $1.15\mathrm{e}{-07} \pm 3.16\mathrm{e}{-08}$ \\

Decision Tree 
& $0.809 \pm 0.098$ & $0.815 \pm 0.106$ 
& $1.15\mathrm{e}{-07} \pm 7.28\mathrm{e}{-08}$ 
& $1.23\mathrm{e}{-07} \pm 5.62\mathrm{e}{-08}$ \\

Random Forest 
& $0.846 \pm 0.062$ & $0.854 \pm 0.069$ 
& $1.05\mathrm{e}{-07} \pm 6.84\mathrm{e}{-08}$ 
& $1.15\mathrm{e}{-07} \pm 5.07\mathrm{e}{-08}$ \\

CNN           
& $0.772 \pm 0.055$ & $0.833 \pm 0.067$ 
& $8.65\mathrm{e}{-08} \pm 3.88\mathrm{e}{-08}$ 
& $9.83\mathrm{e}{-08} \pm 3.88\mathrm{e}{-08}$ \\

CNN--LSTM     
& $0.890 \pm 0.081$ & $0.896 \pm 0.095$ 
& $9.39\mathrm{e}{-08} \pm 4.65\mathrm{e}{-08}$ 
& $1.09\mathrm{e}{-07} \pm 3.95\mathrm{e}{-08}$ \\

DeepAR        
& $0.831 \pm 0.072$ & $0.857 \pm 0.089$ 
& $8.90\mathrm{e}{-08} \pm 3.76\mathrm{e}{-08}$ 
& $1.07\mathrm{e}{-07} \pm 3.49\mathrm{e}{-08}$ \\

N-BEATS       
& $\mathbf{0.732 \pm 0.087}$ & $\mathbf{0.744 \pm 0.125}$ 
& $1.23\mathrm{e}{-07} \pm 8.58\mathrm{e}{-08}$ 
& $1.36\mathrm{e}{-07} \pm 9.12\mathrm{e}{-08}$ \\

Transformer   
& $\mathbf{0.729 \pm 0.071}$ & $\mathbf{0.737 \pm 0.122}$ 
& $9.53\mathrm{e}{-08} \pm 4.91\mathrm{e}{-08}$ 
& $1.07\mathrm{e}{-07} \pm 4.49\mathrm{e}{-08}$ \\

\textbf{PhysAttNet} 
& $\mathbf{0.758 \pm 0.055}$ & $\mathbf{0.765 \pm 0.071}$ 
& $\mathbf{8.19\mathrm{e}{-08} \pm 3.52\mathrm{e}{-08}}$ 
& $\mathbf{9.79\mathrm{e}{-08} \pm 3.43\mathrm{e}{-08}}$ \\
\bottomrule
\end{tabular}}
\end{table}

To assess attention alignment with domain priors, we visualize temporal attention weights alongside observed and predicted signals.


\begin{figure}
    \centering
    \includegraphics[width=0.75\linewidth]{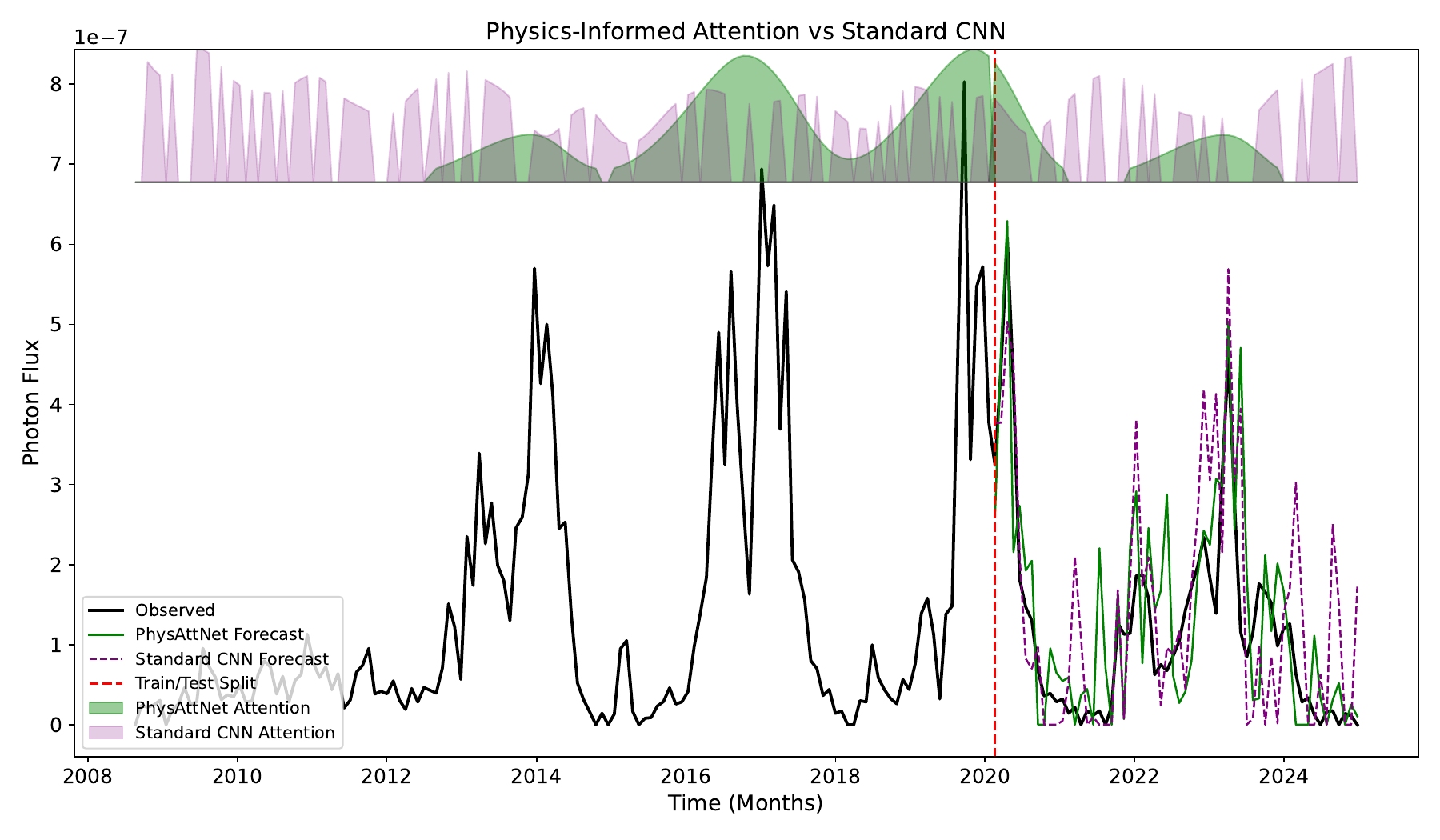}
    \caption{Comparison of performance and attention distribution between PhysAttNet and standard CNN for blazar photon-flux forecasting.}
    \label{fig:blazaratt}
\end{figure}


Figure \ref{fig:blazaratt} compares PhysAttNet and a standard CNN on blazar flare forecasting, showing the full time series with attention over the training period and predictions over the test period.
In the training region, PhysAttNet’s attention concentrates on temporally localized high-flux intervals corresponding to flare-like structures, while remaining relatively suppressed during quiescent phases. In contrast, the standard CNN exhibits more diffuse and less selective attention across time. This indicates that the physics-informed regularization encourages PhysAttNet to prioritize physically meaningful, peak-centered regions rather than distributing importance uniformly.
In the test region, PhysAttNet more accurately tracks both the amplitude and timing of flare events, whereas the standard CNN tends to produce noisy peaks and false extreme values. The improved alignment between attention focus and flare dynamics appears to translate into better event-sensitive forecasting performance.
This shows that guiding attention with domain-informed constraints leads to more selective temporal focus during training and improved predictive performance during testing, particularly around high-activity periods.



\section{Discussion}

The experimental results highlight the potential of physics-informed attention regularization as a transferable inductive bias for time series forecasting across domains governed by physical processes. Although the two application scenarios considered in this study, machining, differ substantially in their spatial and temporal scales, the proposed PhysAttNet framework demonstrates consistent and competitive performance in both settings.

In the blazar forecasting task, PhysAttNet achieves the best overall results across both RMSE and event-weighted RMSE metrics. This improvement is particularly evident during flare periods, where forecasting difficulty is greatest. The attention visualizations further confirm that the model concentrates relevance on temporally localized high-flux intervals corresponding to flare activity. This suggests that the physics-informed attention constraints effectively guide the model toward physically meaningful temporal structures, improving predictive accuracy in regions that are most relevant for scientific interpretation.

For the cutting-force forecasting task, large deep forecasting architectures such as Transformers and N-BEATS achieve the lowest RMSE values. However, PhysAttNet attains comparable performance while relying on a substantially simpler architecture built upon a lightweight CNN backbone. This highlights an important practical advantage: by incorporating domain-informed attention regularization, PhysAttNet can achieve competitive forecasting performance without the computational complexity typically associated with large sequence models. Such efficiency is particularly relevant in industrial monitoring environments where high-frequency signals must be processed in real time.

Taken together, these results suggest that structural assumptions about temporal signals, such as localized event patterns and selective relevance of past observations, can transfer across domains that differ by many orders of magnitude in their physical scale and measurement resolution. In the present study, these principles apply both to gamma-ray light curves observed from distant astrophysical sources and to high-frequency force measurements recorded directly from machining equipment. This cross-domain applicability highlights the potential of physics-informed attention mechanisms as a general modeling paradigm for forecasting in scientific and industrial time series.

At the same time, the current formulation primarily targets signals in which relevant events appear as structured temporal patterns with coherent evolution. In practice, physical systems may also exhibit abrupt or discontinuous dynamics. For example, machining processes can experience sudden chatter onset or rapid engagement changes when cutting complex geometries, while astrophysical sources may produce highly bursty or irregular flare activity. These dynamics may not always conform to the smooth or peak-centered attention patterns encouraged by the present regularization scheme.
Future work will therefore investigate extensions of the proposed framework that explicitly account for such dynamics. Potential directions include adaptive attention regularization that allows localized discontinuities, piecewise attention structures capable of modeling intermittent regimes, or hybrid architectures that combine physics-informed attention with change-point detection mechanisms. Such extensions could further enhance the robustness of forecasting models in environments characterized by highly irregular or rapidly evolving physical processes.
\section{Conclusion}

This work introduced \emph{PhysAttNet}, a physics-informed attention-based CNN architecture for time series forecasting in scientific and industrial applications. Rather than relying solely on increasingly complex neural architectures, the proposed approach incorporates domain-informed structural priors directly into the learning process through attention regularization. These constraints encourage the model to focus on temporally localized and physically meaningful signal regions while maintaining computational efficiency.

The framework was evaluated on two application scenarios with fundamentally different physical characteristics, namely astrophysical flare forecasting from Fermi-LAT observations and high-frequency cutting-force prediction in milling processes. Despite the substantial differences in temporal scale, measurement modality, and signal dynamics, PhysAttNet achieved competitive performance across both domains and demonstrated improved forecasting accuracy in event-dominated regions, particularly in the astrophysical setting. At the same time, the model maintains a lightweight architecture compared to large deep forecasting models such as Transformers and N-BEATS.

These findings suggest that incorporating domain-informed inductive biases into attention mechanisms can provide a principled and efficient alternative to purely architecture-driven approaches in time series forecasting. Future work will explore extensions that account for highly irregular or discontinuous dynamics, such as sudden chatter in machining processes or bursty astrophysical variability, as well as broader applications of physics-informed attention mechanisms in other scientific and industrial time series domains.

\bibliographystyle{cas-model2-names}

\bibliography{literature.bib}



\end{document}